\pdfoutput=1

\documentclass[11pt]{article}

\usepackage[preprint]{acl}

\usepackage{times}
\usepackage{latexsym}
\usepackage{graphicx} 
\usepackage[T1]{fontenc}

\usepackage[utf8]{inputenc}

\usepackage{microtype}
\usepackage{inconsolata}
\usepackage{enumitem}
\usepackage{amssymb}
\usepackage{pifont}
\usepackage{makecell}
\usepackage{booktabs}
\usepackage{adjustbox}
\usepackage{algorithm}
\usepackage{algorithmic}
\usepackage{etoolbox}
\usepackage{multirow}
\usepackage{pifont}
\usepackage[cc]{titlepic}

\newcommand{\vpara}[1]{\vspace{0.03in}\noindent\textbf{#1}}

\newcommand{\cmark}{\ding{51}}%
\newcommand{\xmark}{\ding{55}}%

\title{FreeEval: A Modular Framework for Trustworthy and Efficient Evaluation of Large Language Models}

\author{
    \textbf{
        Zhuohao Yu\textsuperscript{{1}},
        Chang Gao\textsuperscript{{1}},
        Wenjin Yao\textsuperscript{{1}},
        Yidong Wang\textsuperscript{{1}},
        Zhengran Zeng\textsuperscript{{1}},
    } \\
    \textbf{
        Wei Ye\textsuperscript{{1}}\thanks{\llap{}\:Corresponding author. },
        Jindong Wang\textsuperscript{{2}},
        Yue Zhang\textsuperscript{{3}},
        Shikun Zhang\textsuperscript{{1}}
    } \\
    \textsuperscript{1}Peking University.
    \textsuperscript{2}Microsoft Research.
    \textsuperscript{3}Westlake University. \\
    \texttt{zyu@stu.pku.edu.cn, wye@pku.edu.cn}
}

\begin{document}
\maketitle
\begin{abstract}

The rapid development of large language model (LLM) evaluation methodologies and datasets has led to a profound challenge: integrating state-of-the-art evaluation techniques cost-effectively while ensuring reliability, reproducibility, and efficiency. Currently, there is a notable absence of a unified and adaptable framework that seamlessly integrates various evaluation approaches. Moreover, the reliability of evaluation findings is often questionable due to potential data contamination, with the evaluation efficiency commonly overlooked when facing the substantial costs associated with LLM inference. In response to these challenges, we introduce FreeEval, a modular and scalable framework crafted to enable trustworthy and efficient automatic evaluations of LLMs. Firstly, FreeEval's unified abstractions simplify the integration and improve the transparency of diverse evaluation methodologies, encompassing dynamic evaluation  that demand sophisticated LLM interactions. Secondly, the framework integrates meta-evaluation techniques like human evaluation and data contamination detection, which, along with dynamic evaluation modules in the platform, enhance the fairness of the evaluation outcomes. Lastly, FreeEval is designed with a high-performance infrastructure, including distributed computation and caching strategies, enabling extensive evaluations across multi-node, multi-GPU clusters for open-source and proprietary LLMs. We open-source all our code at~\url{https://github.com/WisdomShell/FreeEval}.
\end{abstract}

\section{Introduction}

\begin{table*}
\caption{Comparison of popular evaluation toolkits on features.}
\centering
\small 
\begin{adjustbox}{width=0.9\textwidth}
\label{tab:feature-compare}
\setlength{\tabcolsep}{3pt} 
\renewcommand{\arraystretch}{0.95} 
\begin{tabular}{l|cccccccc}
\toprule
Toolkit &
  \thead{Custom \\ Datasets} &
  \thead{Custom \\ Models} &
  \thead{Custom \\ Prompting} &
  \thead{LLM \\ Judges} &
  \thead{Dynamic \\ Evaluation} &
  \thead{Distributed \\ Inference} &
  \thead{Contamination \\ Detection} &
  \thead{Meta \\ Evaluation} \\ \midrule
\footnotesize Eval-Harness~\citep{eval-harness}      & \cmark & \cmark & \cmark & \xmark & \xmark & \xmark & \xmark & \xmark \\
\footnotesize HELM~\citep{liang2022holistic}         & \cmark & \cmark & \cmark & \xmark & \xmark & \xmark & \xmark & \xmark \\
\footnotesize OpenAI Evals~\citep{openaievals}       & \cmark & \cmark & \cmark & \cmark & \xmark & \xmark & \xmark & \cmark \\
\footnotesize BIG-Bench~\citep{srivastava2023beyond} & \cmark & \cmark & \cmark & \xmark & \xmark & \xmark & \xmark & \xmark \\
\footnotesize OpenCompass~\citep{2023opencompass}    & \cmark & \cmark & \cmark & \cmark & \xmark & \cmark & \xmark & \xmark \\
\footnotesize PromptBench~\citep{zhu2023promptbench} & \cmark & \cmark & \cmark & \xmark & \cmark & \xmark & \xmark & \xmark \\
\footnotesize FreeEval~(Ours)                        & \cmark & \cmark & \cmark & \cmark & \cmark & \cmark & \cmark & \cmark \\ \bottomrule
\end{tabular}
\end{adjustbox}
\end{table*}

Large Language Models (LLMs) have revolutionized the field of Natural Language Processing (NLP) with their impressive performance on a wide range of tasks~\cite{brown2020language, zhang2022opt, bubeck2023sparks, openai2023gpt4}. 
As LLMs play a critical role in both academia and industry, understanding their capabilities and evaluating their performance has become an essential topic~\citep{guo2023evaluating}. 
Therefore,  there have been proposals of automatic evaluation methodologies that harness benchmark datasets~\citep{clark2018arc,zellers2019hellaswag,cobbe2021traininggsm8k,bang2023multitask} for objective assessments, complemented by the introduction of LLM-based subjective evaluation tools~\citep{pandalm,mtbench,alpacaeval,chan2023chateval}.

With the continuous emergence of evaluation data and methods for LLMs, the challenge of incorporating the latest cutting-edge evaluation techniques cost-effectively and conducting more rapid and reliable evaluation, has intensified. In response to this need, several open-source evaluation platforms or toolkits for LLMs were proposed, each with its unique features and focus. \autoref{tab:feature-compare} provides a comprehensive comparison of these frameworks. Specifically, Eval-Harness~\citep{eval-harness} proposes a framework for evaluating LLMs with a variety of benchmark datasets. HELM~\citep{liang2022holistic} provides a collection of metrics beyond accuracy on custom datasets and models. OpenAI Evals~\citep{openaievals} implement interface for creating LLM judges, which leverage LLMs to evaluate other models, and meta-evaluation of these judges. OpenCompass~\citep{2023opencompass} introduces distributed inference with SLURM~\citep{yoo2003slurm} on cluster environments. PromptBench~\citep{zhu2023promptbench} introduces prompt attacks during inference and DyVal~\citep{zhu2023dyval} in the framework. 

Despite these promising efforts, current evaluation platforms still face three bottlenecks. 

First, \emph{a unified and extensible framework is lacking to integrate different evaluation methods seamlessly}. This issue consequently affects evaluation flexibility and transparency. The evaluation results of LLMs may highly depend on complex deployment settings and prompting techniques, since LLMs are not robust enough to handle these intricate settings~\citep{zheng2023large}. For example, \autoref{tab:prompting-compare} demonstrates that these settings can significantly influence results, confirming the need for standardized implementation of evaluation methods to assure transparent and consistent assessment.

Second, \emph{the reliability of empirical results on these platforms can not always be guaranteed}. Automatic evaluation of LLMs remains a complex and challenging task~\citep{chang2023survey} due to their open-ended nature and the presence of data contamination in training datasets, which lead to inflated performance metrics~\citep{schaeffer2023pretraining, sainz2023nlp,yu2024kieval}.

Third, \emph{the efficiency of previous evaluation toolkits is often neglected}. LLM inference might be a significant challenge for researchers since it require strong GPUs or paid APIs, especially when facing large scale evaluations~\citep{pandalm}. Optimizing the inference process and reducing computational costs are crucial for making LLM evaluation more accessible for the research community. 

\begin{table}
\caption{Comparison of different inference implementations. We report 25-shot accuracy of \texttt{llama-2-7b-chat} on ARC-Challenge~\citep{clark2018arc}, 5-shot accuracy on MMLU~\citep{hendrycks2020mmlu} and HellaSwag~\citep{zellers2019hellaswag}. `CP' and `MCP' denote Cloze Prompt and Multiple Choice Prompt from~\citet{Robinson2022LeveragingLL}.}
\centering
\scriptsize
\begin{adjustbox}{width=0.45\textwidth}
\label{tab:prompting-compare}
\setlength{\tabcolsep}{4pt} 
\renewcommand{\arraystretch}{0.9} 
\begin{tabular}{l|ccc}
\toprule
Method     & ARC-C   & MMLU    & HellaSwag \\ \midrule
CP+PromptA  & 51.11\% & 40.65\% & 50.07\%   \\
CP+PromptB  & 47.53\% & 38.72\% & 50.19\%   \\
MCP+PromptA & 54.18\% & 42.73\% & 30.61\%   \\
MCP+PromptB & 54.10\% & 41.28\% & 30.96\%   \\ \bottomrule
\end{tabular}
\end{adjustbox}
\end{table}

To address these challenges, we propose FreeEval, a modular and extensible framework for trustworthy and efficient automatic evaluation of LLMs. The main features of FreeEval are as follows:

\emph{FreeEval offers a unified abstraction and modular implementation of various evaluation methods}. We present the concept of step, dataset, and config to uniformly describe dataset-based, classic reference-based, and LLM-based evaluators. Dataset-based evaluators include task-specific datasets along with dataset operations such as custom prompting, data augmenting, and data generation. LLM-based evaluators, such as MT-Bench~\citep{mtbench}, AlpacaEval~\cite{alpacaeval}, PandaLM~\citep{pandalm} and KIEval~\citep{yu2024kieval}, are also integrated to provide subjective assessment. Complementing these are Classic Judges, which utilize reference-based evaluation metrics like ROUGE~\citep{lin2004rouge} and BERTScore~\citep{zhang2019bertscore} to examine model output. FreeEval's modular design allows for easy implementation of new evaluation protocols and supports evaluating both open-source and proprietary models. The abstractions also bring transparency to the evaluation process since all the evaluation details and settings are open to users.

\emph{FreeEval pioneeringly incorporates several practical meta-evaluation modules to ensure trustworthiness.} Meta-evaluation methods we support include contamination detection, human judgment, case analysis and visualization, and bias evaluation, helping to mitigate the overfitting risks and provide interpretability in model evaluation. FreeEval also includes a user-friendly interface for human annotation to facilitate meta-evaluation and improve the explainability and reliability of evaluation results.

\emph{ FreeEval optimizes distributed and concurrent inference with load balancing and caching mechanisms.} By leveraging cutting-edge inference engines and incorporating well-designed concurrency and caching strategies, FreeEval efficiently handles large-scale evaluations on multi-node multi-GPU clusters. This infrastructure supports a blend of open-source models and proprietary APIs, ensuring scalability and efficient model evaluation while saving inference costs for LLMs.

\section{Background}

In this section, we provide an overview of the current landscape of LLM evaluation methods, the challenges posed by data contamination, and the importance of meta-evaluation in assessing the reliability and validity of evaluation protocols.

\subsection{Automatic Evaluation Methods for LLMs}

The rapid development of Large Language Models (LLMs) has led to the emergence of various evaluation methods, each aiming to assess different aspects of model performance. These methods can be broadly categorized into three groups: classic reference-based evaluation, dataset-based benchmarks, and LLM-based evaluators.

\vpara{Reference-Based Evaluation} methods, such as BLEU~\cite{papineni2002bleu}, ROUGE~\cite{lin2004rouge}, and BERTScore~\cite{zhang2019bertscore}, assess the quality of generated text by comparing it against human-written references. While straightforward, they may not fully capture the open-ended nature of LLM-generated outputs and can be sensitive to reference quality and diversity~\citep{pandalm}.

\vpara{Dataset-Based Benchmarks}, such as ARC \cite{clark2018arc}, HellaSwag~\citep{zellers2019hellaswag}, MMLU \cite{hendrycks2020mmlu}, and C-Eval~\citep{huang2023ceval}, evaluate LLMs using carefully curated datasets that test specific skills or knowledge. However, they may not fully capture the open-ended nature of LLMs and can be vulnerable to data contamination~\cite{schaeffer2023pretraining, wei2023skywork}.

\vpara{LLM-Based Evaluators} leverage strong LLMs, such as GPT-4 \cite{openai2023gpt4}, to assess the performance of other models. Examples include PandaLM~\citep{pandalm}, MT-Bench~\cite{mtbench}, GPTScore~\cite{fu2023gptscore}, PRD~\citep{li2023prd}, and KIEval~\citep{yu2024kieval}. These evaluators can capture nuanced aspects of language understanding and generation, but their performance is influenced by the evaluator LLM and prompting strategies. Biases present in the evaluator LLM may propagate to the evaluation process~\cite{zeng2023evaluating, wang2023large}, requiring careful meta-evaluation. Additionally, the inference cost of LLMs necessitates optimization for large-scale evaluation.

\subsection{Meta-Evaluation of LLMs}
\label{sec:meta-evaluation}
Meta-evaluation refers to the process of evaluating the fairness, reliability, and validity of evaluation protocols themselves. We incorporate several meta-evaluation methods into FreeEval.

\vpara{Data Contamination} occurs when an LLM is exposed to test data during training, leading to inflated performance scores and an inaccurate assessment of the model's true capabilities~\cite{schaeffer2023pretraining, sainz2023nlp, zhu2023dyval}. 
This issue is particularly important due to its impact on evaluation fairness, and should be considered. We implement data contamination detection methods like Min-K prob~\citep{shi2023detecting} and average loss~\citep{wei2023skywork} in FreeEval as modules, to make contamination detection a fundamental process in evaluating LLMs or creating a new evaluation protocol.

\vpara{Human Evaluation} is the gold standard for meta-evaluation~\citep{chang2023survey}, as it directly reflects human preferences on generated texts. This is particularly important for LLM-based evaluators, which subjectively evaluate output quality like human experts. However, the lack of standardized platforms or guidelines for human annotation can lead to biased, inconsistent, and unfair judgments. To address this, we incorporate meta-evaluation protocols from~\citet{pandalm, zeng2023evaluating, mtbench}, as they reflect preferences from human experts in different scenarios. Additionally, we create a user-friendly interface for human experts to create new preference datasets, facilitating the collection of high-quality human evaluations for meta-evaluation purposes.

\begin{figure*}[t!]
	\centering
	\includegraphics[width=1.0\textwidth]{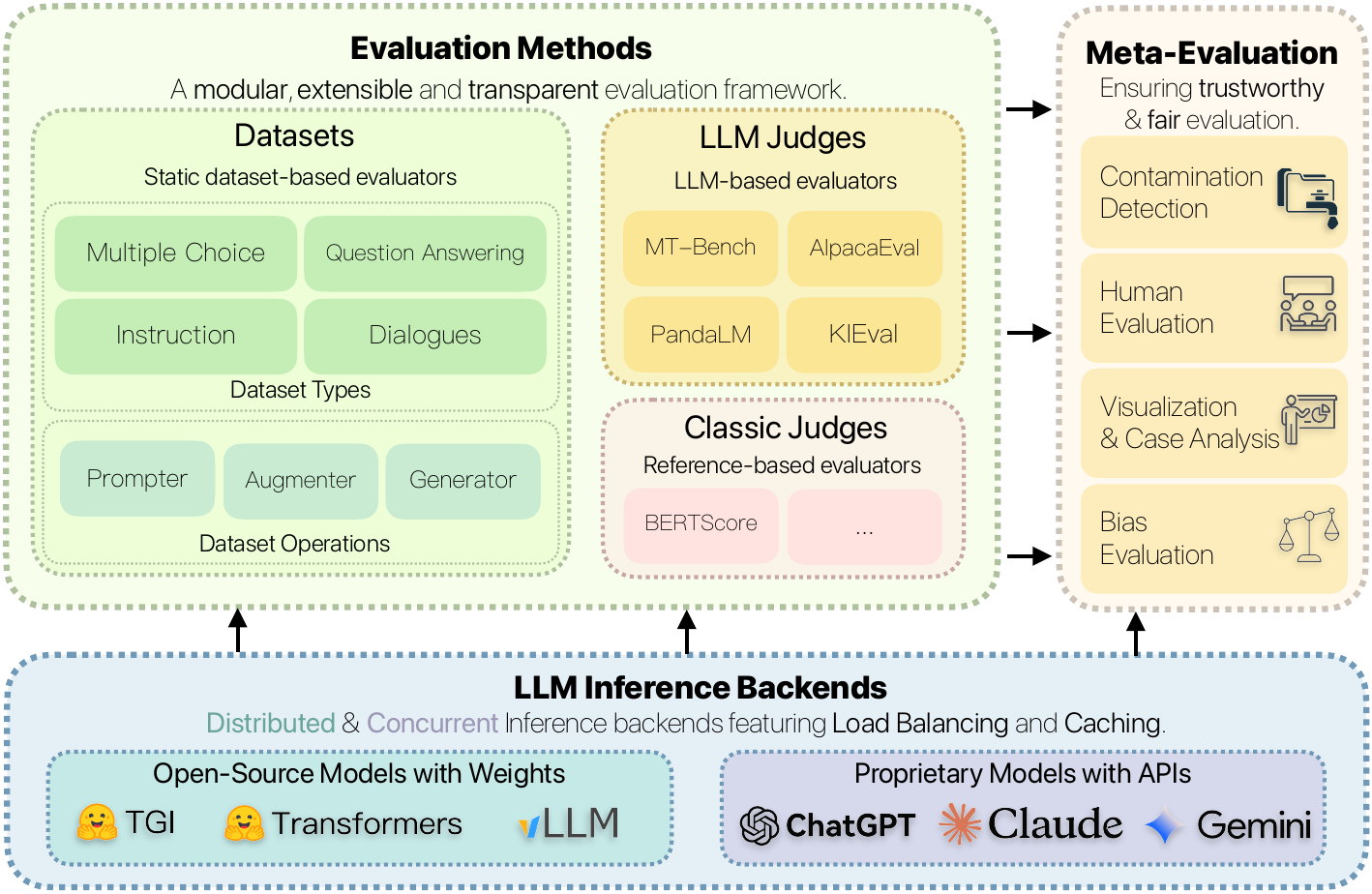}
	\caption{Overall architecture of FreeEval.}
	\label{fig-pipeline}
\end{figure*}

\section{Design and Implementation}

In this section, we present the design and implementation of FreeEval, we discuss the framework's architecture, its key components, and how they address the challenges identified previously.

\subsection{Design Principles}

To build a flexible, efficient research tool for LLM evaluation we make sure the architecture of FreeEval follows the following principles:
\begin{itemize}[leftmargin=0.5em]
	\setlength\itemsep{0.1em}
     \item \textbf{Modular}: FreeEval provides a modular architecture that allows for easy integration of new evaluation methods, datasets, and protocols. This modularity also ensures transparency by making all evaluation settings and details openly accessible to users.

    \item \textbf{Trustworthy}: The evaluation results must be trustworthy, and the evaluation process should be fair and effective. FreeEval allows users to propose new evaluation methods, with a comprehensive meta-evaluation proving its soundness.
    
    \item \textbf{Efficient}: FreeEval prioritizes efficiency to minimize the high computational costs associated with LLM inference. By focusing on cost-effective evaluation processes, researchers can conduct large-scale evaluations while effectively managing computational resources and financial costs. 
\end{itemize}

\subsection{FreeEval Architecture Overview}

FreeEval's architecture, illustrated in Figure~\ref{fig-pipeline}, features a modular design that could be separated into Evaluation Methods, Meta-Evaluation and LLM Inference Backends. Evaluation Methods contain different datasets and implementation for evaluation methods. The Meta-Evaluation module ensures the integrity and fairness of assessments by providing data contamination detection methods and popular meta-evaluation method implementation. LLM Inference Backends form the computational backbone, as it provide distributed and concurrent inference of LLMs featuring performance optimization techniques.

\subsection{Extensible Modular Design}

FreeEval's modular architecture is designed to accommodate the rapidly evolving landscape of LLM evaluation.
To help users implement new evaluation methods without complexity, FreeEval is implemented around the concept of \texttt{step}, \texttt{dataset} and \texttt{config}, which serve as the building blocks for creating flexible and extensible evaluation pipelines:

\begin{itemize}[leftmargin=0.5em]
\setlength\itemsep{0.1em}
    \item \textbf{\texttt{step}}: A \texttt{step} encapsulates a specific evaluation method, data augmentation technique, or metric calculation logic. Each step contain three phases: \texttt{preprocess} handles loading or initializing the required \texttt{dataset} or models; \texttt{run} handles the execution of actual logics; \texttt{postprocess} parse the outputs, collects evaluation results and free up the resources.

    \item \textbf{\texttt{dataset}}: Data used by the evaluators are defined as \texttt{dataset}. Each \texttt{dataset} handles the pre-processing required to load data, few-shot settings, prompting, augmentation of instances, and post-processing of inference results.
    
    \item \textbf{\texttt{config}}: A \texttt{config} file is used to compose evaluation pipelines with \texttt{step}s and \texttt{dataset}s. The config file contains all the details and settings. \texttt{step}s defined in the \texttt{config} are executed sequentially, and they share the same context which stores intermediate results.
\end{itemize}

These abstractions improve transparency in evaluations by providing users with full access to the configuration details for each evaluation pipeline. The \texttt{config} file also serves as a complete record of the evaluation process, including all necessary hyperparameters and settings. The modular design also allow data to be re-used in different scenarios without effort. For example, GSM8K~\citep{cobbe2021traininggsm8k} is a evaluation dataset, we could simply calculate perplexity of models on this dataset, or we could use a data generation \texttt{step} to generate new data with GPT-4 in the same distribution to detect data contamination following~\citet{wei2023skywork}. The modular approach  allows researchers to easily add new evaluation methods or modify existing ones without disrupting the overall structure of the framework. By defining each evaluator as a self-contained unit, FreeEval promotes code reusability and maintainability.

\begin{figure}
    \centering
    \includegraphics[width=0.45\textwidth]{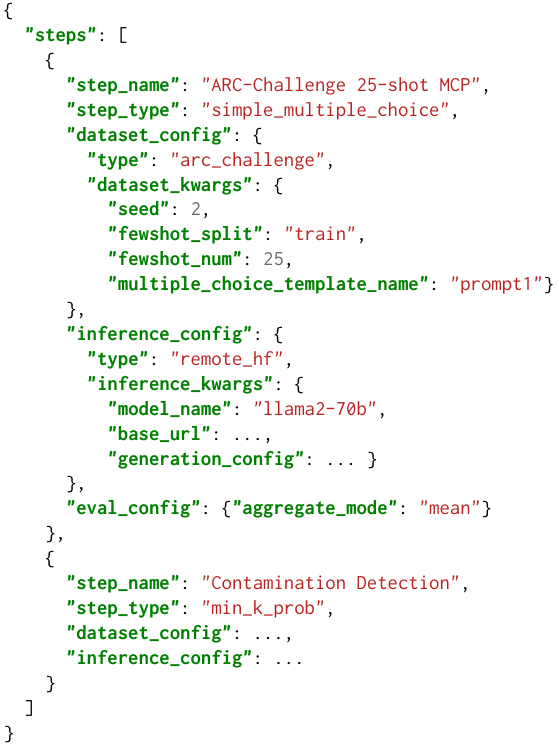}
    \caption{Config for an example pipeline, evaluating LLaMA-2 70B~\citep{touvron2023llama2} on ARC-Challenge~\citep{clark2018arc} dataset and then detecting data contamination with Min-K\% Prob~\citep{shi2023detecting}.}
    \label{fig:config-example}
\end{figure}

This configuration-driven approach eliminates the need for users to write Python code when defining and running an evaluation pipeline. All settings and parameters for each \texttt{step} and \texttt{dataset} are specified within the \texttt{config} file, making the evaluation process highly customizable and accessible to researchers with varying levels of programming expertise.~\autoref{fig:config-example} shows an example config for a pipeline evaluating LLaMA-2 70B~\citep{touvron2023llama2} on ARC-Challenge~\citep{clark2018arc} dataset with a fixed seed for sampling 25-shot examples and custom prompt. The model can be deployed locally or on a remote machine. The pipeline also include detecting data contamination with Min-K\% Prob~\citep{shi2023detecting}.

\subsection{Trustworthy Evaluation}

FreeEval prioritizes trustworthiness and fairness in evaluations by incorporating a range of meta-evaluation modules that validates the evaluation results and processes. 

As human preference remain the gold standard for measuring the effectiveness of evaluation protocols, FreeEval model human preference into two types: \emph{pairwise comparison} and \emph{direct scoring}. We incorporate existing meta-evaluation datasets from PandaLM~\citep{pandalm}, MT-Bench~\citep{mtbench}, LLMBar~\citep{guo2023evaluating}, AlpacaEval~\citep{alpacaeval}, and provide a user-friendly interface for annotating and curating new human evaluation datasets. 

To ensure the trustworthiness of the evaluation results, we also implement data contamination detection methods, as introduced in ~\autoref{sec:meta-evaluation}, into our toolkit as \texttt{step}s. Understanding whether the tested dataset appear in the training phase of the evaluated models would help users assess the validity and reliability of evaluation results. We also provide bias evaluation modules and visualization tools specifically for LLM-based evaluators, as previous studies have reported they exhibit position bias and length bias~\citep{mtbench, pandalm}. These meta-evaluation modules can be easily integrated into existing evaluation pipelines, allowing researchers to understand the effectiveness of their results, the fairness of the evaluation process, and study bad cases that lead to unexpected evaluation results.

\subsection{Efficient Inference Backends}

\begin{figure}
    \centering
    \includegraphics[width=0.5\textwidth]{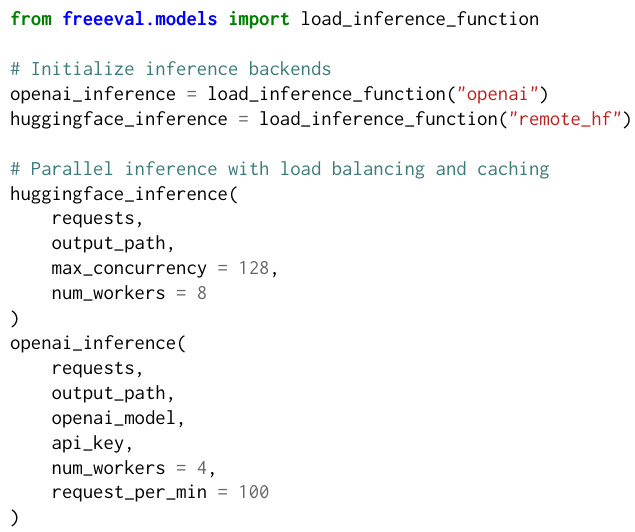}
    \caption{Example code for running FreeEval's inference backends. We rely on these backends for efficient inference and provide a simple abstraction.}
    \label{fig:inference-code}
\end{figure}

FreeEval's high-performance inference backends are designed to efficiently handle the computational demands of large-scale LLM evaluations. 

The inference backends in FreeEval support both open-source models and proprietary models with APIs, providing researchers with flexibility in choosing the LLMs they wish to evaluate. 
For all models, FreeEval support concurrent inference given a fixed number of workers. We implement a caching mechanism for queries based on hash values of the request. We hash the request prompt and inference config, and store locally the request content and response for each individual request. By checking the cache before making a query, FreeEval skips cached requests, enabling quick recovery from exceptions and saving inference costs. This is particularly beneficial when implementing and debugging new evaluation methods. Caching also ensures reproducibility, as all requests, parameters, and responses are saved and can be inspected using FreeEval's visualization tools.

For open-source models, we leverage Huggingface's \texttt{text-generation-inference} (TGI)~\citep{huggingface2023textgeneration} package which is a production-ready high-performance inference toolkit. We implement a load-balancing technique in conjunction with the continuous batching feature provided by TGI to maximize GPU utilization on multi-node multi-GPU clusters. For proprietary models, we also implement a rate-limiting mechanism so that users could define their total number of requests per minute, to avoid causing too much stress on API providers. 

We evaluate FreeEval's performance by comparing the execution times (excluding dataset downloading times) for \texttt{llama-2-7b-chat-hf} model on 3 common datasets using different toolkits. Our experiments are done on the same Ubuntu machine with a single NVIDIA A800 80GB PCIe GPU and Intel Xeon Gold CPU. As shown in~\autoref{tab:performance}, even on a single GPU, FreeEval (with concurrent execution enabled) exhibit significant advantage on all benchmark datasets.

The inference backends in FreeEval are designed to seamlessly integrate with the evaluation methods and meta-evaluation components of the framework. As illustrated in \autoref{fig:inference-code}, initializing the inference backends and running parallel inference is straightforward and user-friendly. This simplicity allows developers of new evaluation methods to focus on prompting or interactions between models, using the backends sequentially. As a result, implementing interactive evaluation methods, such as those proposed by \citet{li2023prd, chan2023chateval, yu2024kieval}, becomes much easier and more accessible to researchers.

\begin{table}
\caption{Comparison of execution time (in hours) of different toolkits. All experiments are done on the same machine with a single NVIDIA A800 80GB PCIe GPU.}
\centering
\scriptsize
\begin{adjustbox}{width=0.45\textwidth}
\label{tab:performance}
\setlength{\tabcolsep}{4pt} 
\renewcommand{\arraystretch}{0.95} 
\begin{tabular}{l|ccc}
\toprule
Toolkit                      & ARC-C          & MMLU           & HellaSwag      \\ \midrule
Eval-Harness          & 0.160          &  0.510         & 1.080          \\
OpenCompass           &   0.084   & 1.431          & 1.716          \\
FreeEval (Sequential) & 0.211          & 0.949          & 0.966    \\
FreeEval (Concurrent) & \textbf{0.067} & \textbf{0.233} & \textbf{0.357} \\     \bottomrule
\end{tabular}
\end{adjustbox}
\end{table}
\section{Conclusion}

In this paper, we introduce FreeEval, a modular and extensible framework for trustworthy and efficient automatic evaluation of LLMs. FreeEval addresses the challenges of standardization, reliability, and efficiency in LLM evaluation by providing a unified implementation of various evaluation methods, incorporating meta-evaluation modules, and leveraging high-performance inference backends. The framework's modular design allows for easy integration of new evaluation protocols and improves transparency for evaluation methods. By focusing on meta-evaluation, reproducibility, and efficiency, FreeEval aims to promote the development of more reliable and fair evaluation methods for LLMs, contributing to the development of more robust and trustworthy language models. 
In the future, we will continuously maintain the FreeEval toolkit and we hope FreeEval will serve as a useful tool for researchers and practitioners to gain deeper insights into the capabilities and limitations of LLMs.

\appendix

\section{Ethical Conciderations}

The development and evaluation of LLMs raise several ethical concerns that must be carefully considered. FreeEval aims to address some of these concerns by promoting transparency, fairness, and reliability in LLM evaluation. However, it is important to acknowledge that the framework itself does not eliminate all ethical risks associated with LLMs.

One key ethical consideration is the potential for biases and discrimination in LLMs. FreeEval incorporates bias evaluation modules to help identify and mitigate biases in the evaluation process, but it cannot fully eliminate biases present in the training data or the models themselves. Researchers using FreeEval should be aware of these limitations and strive to develop more inclusive and equitable LLMs.

Another ethical concern is the environmental impact of training and evaluating LLMs, which can require significant computational resources and energy consumption. FreeEval's efficient inference backends help reduce the computational costs of evaluation, but the overall environmental impact of LLM development remains a pressing issue that requires further research and innovation.

Finally, the use of LLMs in real-world applications raises questions about accountability, transparency, and the potential for misuse. While FreeEval promotes transparency in the evaluation process, it is ultimately the responsibility of researchers and developers to ensure that LLMs are deployed ethically and with appropriate safeguards in place.

By providing a standardized and comprehensive evaluation framework, FreeEval aims to contribute to the ongoing dialogue around the ethical implications of LLMs and support the development of more responsible and trustworthy language models.

\nocite{*}
\bibliography{custom}

\begin{thebibliography}{106}
\expandafter\ifx\csname natexlab\endcsname\relax\def\natexlab#1{#1}\fi

\bibitem[{Anil et~al.(2023)Anil, Dai, Firat, Johnson, Lepikhin, Passos,
  Shakeri, Taropa, Bailey, Chen et~al.}]{anil2023palm}
Rohan Anil, Andrew~M Dai, Orhan Firat, Melvin Johnson, Dmitry Lepikhin,
  Alexandre Passos, Siamak Shakeri, Emanuel Taropa, Paige Bailey, Zhifeng Chen,
  et~al. 2023.
\newblock Palm 2 technical report.
\newblock \emph{arXiv preprint arXiv:2305.10403}.

\bibitem[{Arpit et~al.(2017)Arpit, Jastrz{\k{e}}bski, Ballas, Krueger, Bengio,
  Kanwal, Maharaj, Fischer, Courville, Bengio et~al.}]{arpit2017closer}
Devansh Arpit, Stanis{\l}aw Jastrz{\k{e}}bski, Nicolas Ballas, David Krueger,
  Emmanuel Bengio, Maxinder~S Kanwal, Tegan Maharaj, Asja Fischer, Aaron
  Courville, Yoshua Bengio, et~al. 2017.
\newblock A closer look at memorization in deep networks.
\newblock In \emph{International conference on machine learning}, pages
  233--242. PMLR.

\bibitem[{Banerjee and Lavie(2005)}]{banerjee2005meteor}
Satanjeev Banerjee and Alon Lavie. 2005.
\newblock Meteor: An automatic metric for mt evaluation with improved
  correlation with human judgments.
\newblock In \emph{Proceedings of the acl workshop on intrinsic and extrinsic
  evaluation measures for machine translation and/or summarization}, pages
  65--72.

\bibitem[{Bang et~al.(2023)Bang, Cahyawijaya, Lee, Dai, Su, Wilie, Lovenia, Ji,
  Yu, Chung et~al.}]{bang2023multitask}
Yejin Bang, Samuel Cahyawijaya, Nayeon Lee, Wenliang Dai, Dan Su, Bryan Wilie,
  Holy Lovenia, Ziwei Ji, Tiezheng Yu, Willy Chung, et~al. 2023.
\newblock A multitask, multilingual, multimodal evaluation of chatgpt on
  reasoning, hallucination, and interactivity.
\newblock \emph{arXiv preprint arXiv:2302.04023}.

\bibitem[{Beeching et~al.(2023)Beeching, Fourrier, Habib, Han, Lambert, Rajani,
  Sanseviero, Tunstall, and Wolf}]{open-llm-leaderboard}
Edward Beeching, Clémentine Fourrier, Nathan Habib, Sheon Han, Nathan Lambert,
  Nazneen Rajani, Omar Sanseviero, Lewis Tunstall, and Thomas Wolf. 2023.
\newblock Open llm leaderboard.
\newblock
  \url{https://huggingface.co/spaces/HuggingFaceH4/open_llm_leaderboard}.

\bibitem[{Bengio and LeCun(2007)}]{Bengio+chapter2007}
Yoshua Bengio and Yann LeCun. 2007.
\newblock Scaling learning algorithms towards {AI}.
\newblock In \emph{Large Scale Kernel Machines}. MIT Press.

\bibitem[{Berglund et~al.(2023)Berglund, Tong, Kaufmann, Balesni, Stickland,
  Korbak, and Evans}]{berglund2023reversal}
Lukas Berglund, Meg Tong, Max Kaufmann, Mikita Balesni, Asa~Cooper Stickland,
  Tomasz Korbak, and Owain Evans. 2023.
\newblock The reversal curse: Llms trained on" a is b" fail to learn" b is a".
\newblock \emph{arXiv preprint arXiv:2309.12288}.

\bibitem[{Biderman et~al.(2023{\natexlab{a}})Biderman, Prashanth, Sutawika,
  Schoelkopf, Anthony, Purohit, and Raf}]{biderman2023emergent}
Stella Biderman, USVSN~Sai Prashanth, Lintang Sutawika, Hailey Schoelkopf,
  Quentin Anthony, Shivanshu Purohit, and Edward Raf. 2023{\natexlab{a}}.
\newblock Emergent and predictable memorization in large language models.
\newblock \emph{arXiv preprint arXiv:2304.11158}.

\bibitem[{Biderman et~al.(2023{\natexlab{b}})Biderman, Schoelkopf, Anthony,
  Bradley, O'Brien, Hallahan, Khan, Purohit, Prashanth, Raff
  et~al.}]{biderman2023pythia}
Stella Biderman, Hailey Schoelkopf, Quentin Anthony, Herbie Bradley, Kyle
  O'Brien, Eric Hallahan, Mohammad~Aflah Khan, Shivanshu Purohit, USVSN~Sai
  Prashanth, Edward Raff, et~al. 2023{\natexlab{b}}.
\newblock Pythia: A suite for analyzing large language models across training
  and scaling.
\newblock \emph{arXiv preprint arXiv:2304.01373}.

\bibitem[{Bommasani et~al.(2023)Bommasani, Liang, and
  Lee}]{bommasani2023holistic}
Rishi Bommasani, Percy Liang, and Tony Lee. 2023.
\newblock Holistic evaluation of language models.
\newblock \emph{Annals of the New York Academy of Sciences}.

\bibitem[{Brown et~al.(2020)Brown, Mann, Ryder, Subbiah, Kaplan, Dhariwal,
  Neelakantan, Shyam, Sastry, Askell et~al.}]{brown2020language}
Tom Brown, Benjamin Mann, Nick Ryder, Melanie Subbiah, Jared~D Kaplan, Prafulla
  Dhariwal, Arvind Neelakantan, Pranav Shyam, Girish Sastry, Amanda Askell,
  et~al. 2020.
\newblock Language models are few-shot learners.
\newblock \emph{Advances in neural information processing systems},
  33:1877--1901.

\bibitem[{Bubeck et~al.(2023)Bubeck, Chandrasekaran, Eldan, Gehrke, Horvitz,
  Kamar, Lee, Lee, Li, Lundberg et~al.}]{bubeck2023sparks}
S{\'e}bastien Bubeck, Varun Chandrasekaran, Ronen Eldan, Johannes Gehrke, Eric
  Horvitz, Ece Kamar, Peter Lee, Yin~Tat Lee, Yuanzhi Li, Scott Lundberg,
  et~al. 2023.
\newblock Sparks of artificial general intelligence: Early experiments with
  gpt-4.
\newblock \emph{arXiv preprint arXiv:2303.12712}.

\bibitem[{Chan et~al.(2023)Chan, Chen, Su, Yu, Xue, Zhang, Fu, and
  Liu}]{chan2023chateval}
Chi-Min Chan, Weize Chen, Yusheng Su, Jianxuan Yu, Wei Xue, Shanghang Zhang,
  Jie Fu, and Zhiyuan Liu. 2023.
\newblock Chateval: Towards better llm-based evaluators through multi-agent
  debate.
\newblock \emph{arXiv preprint arXiv:2308.07201}.

\bibitem[{Chang et~al.(2023)Chang, Wang, Wang, Wu, Zhu, Chen, Yang, Yi, Wang,
  Wang et~al.}]{chang2023survey}
Yupeng Chang, Xu~Wang, Jindong Wang, Yuan Wu, Kaijie Zhu, Hao Chen, Linyi Yang,
  Xiaoyuan Yi, Cunxiang Wang, Yidong Wang, et~al. 2023.
\newblock A survey on evaluation of large language models.
\newblock \emph{arXiv preprint arXiv:2307.03109}.

\bibitem[{Chiang and Lee(2023)}]{chiang2023can}
Cheng-Han Chiang and Hung-yi Lee. 2023.
\newblock Can large language models be an alternative to human evaluations?
\newblock \emph{arXiv preprint arXiv:2305.01937}.

\bibitem[{Chiang et~al.(2023)Chiang, Li, Lin, Sheng, Wu, Zhang, Zheng, Zhuang,
  Zhuang, Gonzalez et~al.}]{chiang2023vicuna}
Wei-Lin Chiang, Zhuohan Li, Zi~Lin, Ying Sheng, Zhanghao Wu, Hao Zhang, Lianmin
  Zheng, Siyuan Zhuang, Yonghao Zhuang, Joseph~E Gonzalez, et~al. 2023.
\newblock Vicuna: An open-source chatbot impressing gpt-4 with 90\%* chatgpt
  quality.
\newblock \emph{See https://vicuna. lmsys. org (accessed 14 April 2023)}.

\bibitem[{Clark et~al.(2018)Clark, Cowhey, Etzioni, Khot, Sabharwal, Schoenick,
  and Tafjord}]{clark2018arc}
Peter Clark, Isaac Cowhey, Oren Etzioni, Tushar Khot, Ashish Sabharwal, Carissa
  Schoenick, and Oyvind Tafjord. 2018.
\newblock Think you have solved question answering? try arc, the ai2 reasoning
  challenge.
\newblock \emph{arXiv preprint arXiv:1803.05457}.

\bibitem[{Cobbe et~al.(2021)Cobbe, Kosaraju, Bavarian, Chen, Jun, Kaiser,
  Plappert, Tworek, Hilton, Nakano et~al.}]{cobbe2021traininggsm8k}
Karl Cobbe, Vineet Kosaraju, Mohammad Bavarian, Mark Chen, Heewoo Jun, Lukasz
  Kaiser, Matthias Plappert, Jerry Tworek, Jacob Hilton, Reiichiro Nakano,
  et~al. 2021.
\newblock Training verifiers to solve math word problems.
\newblock \emph{arXiv preprint arXiv:2110.14168}.

\bibitem[{Contributors(2023)}]{srivastava2023beyond}
Contributors. 2023.
\newblock \href {https://openreview.net/forum?id=uyTL5Bvosj} {Beyond the
  imitation game: Quantifying and extrapolating the capabilities of language
  models}.
\newblock \emph{Transactions on Machine Learning Research}.

\bibitem[{{Contributors}(2023)}]{openaievals}
{Contributors}. 2023.
\newblock Openai evals.
\newblock \url{https://github.com/openai/evals}.

\bibitem[{Contributors(2023{\natexlab{a}})}]{huggingface2023textgeneration}
Contributors. 2023{\natexlab{a}}.
\newblock Text generation inference: A rust, python and grpc server for text
  generation inference.
\newblock \url{https://github.com/huggingface/text-generation-inference}.

\bibitem[{Contributors(2023{\natexlab{b}})}]{2023opencompass}
OpenCompass Contributors. 2023{\natexlab{b}}.
\newblock Opencompass: A universal evaluation platform for foundation models.
\newblock \url{https://github.com/open-compass/opencompass}.

\bibitem[{Daniele and Suphavadeeprasit(2023)}]{daniele2023amplify-instruct}
Luigi Daniele and Suphavadeeprasit. 2023.
\newblock Amplify-instruct: Synthetically generated diverse multi-turn
  conversations for effecient llm training.
\newblock \emph{arXiv preprint arXiv:(comming soon)}.

\bibitem[{Dey et~al.(2023)Dey, Gosal, Khachane, Marshall, Pathria, Tom,
  Hestness et~al.}]{dey2023cerebras}
Nolan Dey, Gurpreet Gosal, Hemant Khachane, William Marshall, Ribhu Pathria,
  Marvin Tom, Joel Hestness, et~al. 2023.
\newblock Cerebras-gpt: Open compute-optimal language models trained on the
  cerebras wafer-scale cluster.
\newblock \emph{arXiv preprint arXiv:2304.03208}.

\bibitem[{Dodge et~al.(2020)Dodge, Ilharco, Schwartz, Farhadi, Hajishirzi, and
  Smith}]{dodge2020fine}
Jesse Dodge, Gabriel Ilharco, Roy Schwartz, Ali Farhadi, Hannaneh Hajishirzi,
  and Noah Smith. 2020.
\newblock Fine-tuning pretrained language models: Weight initializations, data
  orders, and early stopping.
\newblock \emph{arXiv preprint arXiv:2002.06305}.

\bibitem[{Du et~al.(2023)Du, Mukherjee, Cheng, Shokouhi, Hu, and
  Hassan}]{du2023robustness}
Mengnan Du, Subhabrata Mukherjee, Yu~Cheng, Milad Shokouhi, Xia Hu, and Ahmed
  Hassan. 2023.
\newblock Robustness challenges in model distillation and pruning for natural
  language understanding.
\newblock In \emph{Proceedings of the 17th Conference of the European Chapter
  of the Association for Computational Linguistics}, pages 1758--1770.

\bibitem[{Duan et~al.(2024)Duan, Suri, Mireshghallah, Min, Shi, Zettlemoyer,
  Tsvetkov, Choi, Evans, and Hajishirzi}]{duan2024membership}
Michael Duan, Anshuman Suri, Niloofar Mireshghallah, Sewon Min, Weijia Shi,
  Luke Zettlemoyer, Yulia Tsvetkov, Yejin Choi, David Evans, and Hannaneh
  Hajishirzi. 2024.
\newblock Do membership inference attacks work on large language models?
\newblock \emph{arXiv preprint arXiv:2402.07841}.

\bibitem[{Dubois et~al.(2023)Dubois, Li, Taori, Zhang, Gulrajani, Ba, Guestrin,
  Liang, and Hashimoto}]{dubois2023alpacafarm}
Yann Dubois, Xuechen Li, Rohan Taori, Tianyi Zhang, Ishaan Gulrajani, Jimmy Ba,
  Carlos Guestrin, Percy Liang, and Tatsunori~B Hashimoto. 2023.
\newblock Alpacafarm: A simulation framework for methods that learn from human
  feedback.
\newblock \emph{arXiv preprint arXiv:2305.14387}.

\bibitem[{Eccleston(2023)}]{sharegpt}
Dom Eccleston. 2023.
\newblock Sharegpt dataset.
\newblock \url{https://sharegpt.com/}.

\bibitem[{Fu et~al.(2023)Fu, Ng, Jiang, and Liu}]{fu2023gptscore}
Jinlan Fu, See-Kiong Ng, Zhengbao Jiang, and Pengfei Liu. 2023.
\newblock Gptscore: Evaluate as you desire.
\newblock \emph{arXiv preprint arXiv:2302.04166}.

\bibitem[{Gao et~al.(2021)Gao, Tow, Biderman, Black, DiPofi, Foster, Golding,
  Hsu, McDonell, Muennighoff, Phang, Reynolds, Tang, Thite, Wang, Wang, and
  Zou}]{eval-harness}
Leo Gao, Jonathan Tow, Stella Biderman, Sid Black, Anthony DiPofi, Charles
  Foster, Laurence Golding, Jeffrey Hsu, Kyle McDonell, Niklas Muennighoff,
  Jason Phang, Laria Reynolds, Eric Tang, Anish Thite, Ben Wang, Kevin Wang,
  and Andy Zou. 2021.
\newblock \href {https://doi.org/10.5281/zenodo.5371628} {A framework for
  few-shot language model evaluation}.

\bibitem[{Godbole et~al.(2023)Godbole, Dahl, Gilmer, Shallue, and
  Nado}]{tuningplaybookgithub}
Varun Godbole, George~E. Dahl, Justin Gilmer, Christopher~J. Shallue, and
  Zachary Nado. 2023.
\newblock \href {http://github.com/google-research/tuning_playbook} {Deep
  learning tuning playbook}.
\newblock Version 1.0.

\bibitem[{Goodfellow et~al.(2016)Goodfellow, Bengio, Courville, and
  Bengio}]{goodfellow2016deep}
Ian Goodfellow, Yoshua Bengio, Aaron Courville, and Yoshua Bengio. 2016.
\newblock \emph{Deep learning}, volume~1.
\newblock MIT Press.

\bibitem[{Google(2023)}]{google2023bard}
Google. 2023.
\newblock Bard.

\bibitem[{Guo et~al.(2023)Guo, Jin, Liu, Huang, Shi, Yu, Liu, Li, Xiong, Xiong
  et~al.}]{guo2023evaluating}
Zishan Guo, Renren Jin, Chuang Liu, Yufei Huang, Dan Shi, Linhao Yu, Yan Liu,
  Jiaxuan Li, Bojian Xiong, Deyi Xiong, et~al. 2023.
\newblock Evaluating large language models: A comprehensive survey.
\newblock \emph{arXiv preprint arXiv:2310.19736}.

\bibitem[{Hendrycks et~al.(2020)Hendrycks, Burns, Basart, Zou, Mazeika, Song,
  and Steinhardt}]{hendrycks2020mmlu}
Dan Hendrycks, Collin Burns, Steven Basart, Andy Zou, Mantas Mazeika, Dawn
  Song, and Jacob Steinhardt. 2020.
\newblock Measuring massive multitask language understanding.
\newblock \emph{arXiv preprint arXiv:2009.03300}.

\bibitem[{Hinton et~al.(2006)Hinton, Osindero, and Teh}]{Hinton06}
Geoffrey~E. Hinton, Simon Osindero, and Yee~Whye Teh. 2006.
\newblock A fast learning algorithm for deep belief nets.
\newblock \emph{Neural Computation}, 18:1527--1554.

\bibitem[{Hirschman and Gaizauskas(2001)}]{hirschman2001natural}
Lynette Hirschman and Robert Gaizauskas. 2001.
\newblock Natural language question answering: the view from here.
\newblock \emph{natural language engineering}, 7(4):275--300.

\bibitem[{Hochreiter and Schmidhuber(1997)}]{hochreiter1997long}
Sepp Hochreiter and J{\"u}rgen Schmidhuber. 1997.
\newblock Long short-term memory.
\newblock \emph{Neural computation}, 9(8):1735--1780.

\bibitem[{Hu et~al.(2021)Hu, Wallis, Allen-Zhu, Li, Wang, Wang, Chen
  et~al.}]{hulora}
Edward~J Hu, Phillip Wallis, Zeyuan Allen-Zhu, Yuanzhi Li, Shean Wang, Lu~Wang,
  Weizhu Chen, et~al. 2021.
\newblock Lora: Low-rank adaptation of large language models.
\newblock In \emph{International Conference on Learning Representations}.

\bibitem[{Huang et~al.(2023)Huang, Bai, Zhu, Zhang, Zhang, Su, Liu, Lv, Zhang,
  Lei et~al.}]{huang2023ceval}
Yuzhen Huang, Yuzhuo Bai, Zhihao Zhu, Junlei Zhang, Jinghan Zhang, Tangjun Su,
  Junteng Liu, Chuancheng Lv, Yikai Zhang, Jiayi Lei, et~al. 2023.
\newblock C-eval: A multi-level multi-discipline chinese evaluation suite for
  foundation models.
\newblock \emph{arXiv preprint arXiv:2305.08322}.

\bibitem[{Jiang et~al.(2023)Jiang, Sablayrolles, Mensch, Bamford, Chaplot,
  Casas, Bressand, Lengyel, Lample, Saulnier et~al.}]{jiang2023mistral}
Albert~Q Jiang, Alexandre Sablayrolles, Arthur Mensch, Chris Bamford,
  Devendra~Singh Chaplot, Diego de~las Casas, Florian Bressand, Gianna Lengyel,
  Guillaume Lample, Lucile Saulnier, et~al. 2023.
\newblock Mistral 7b.
\newblock \emph{arXiv preprint arXiv:2310.06825}.

\bibitem[{Karpinska et~al.(2021)Karpinska, Akoury, and
  Iyyer}]{karpinska2021perils}
Marzena Karpinska, Nader Akoury, and Mohit Iyyer. 2021.
\newblock The perils of using mechanical turk to evaluate open-ended text
  generation.
\newblock \emph{arXiv preprint arXiv:2109.06835}.

\bibitem[{Kenton and Toutanova(2019)}]{kenton2019bert}
Jacob Devlin Ming-Wei~Chang Kenton and Lee~Kristina Toutanova. 2019.
\newblock Bert: Pre-training of deep bidirectional transformers for language
  understanding.
\newblock In \emph{Proceedings of NAACL-HLT}, pages 4171--4186.

\bibitem[{Kwiatkowski et~al.(2019)Kwiatkowski, Palomaki, Redfield, Collins,
  Parikh, Alberti, Epstein, Polosukhin, Devlin, Lee
  et~al.}]{kwiatkowski2019natural}
Tom Kwiatkowski, Jennimaria Palomaki, Olivia Redfield, Michael Collins, Ankur
  Parikh, Chris Alberti, Danielle Epstein, Illia Polosukhin, Jacob Devlin,
  Kenton Lee, et~al. 2019.
\newblock Natural questions: a benchmark for question answering research.
\newblock \emph{Transactions of the Association for Computational Linguistics},
  7:453--466.

\bibitem[{Li et~al.(2023{\natexlab{a}})Li, Patel, and Du}]{li2023prd}
Ruosen Li, Teerth Patel, and Xinya Du. 2023{\natexlab{a}}.
\newblock Prd: Peer rank and discussion improve large language model based
  evaluations.
\newblock \emph{arXiv preprint arXiv:2307.02762}.

\bibitem[{Li et~al.(2023{\natexlab{b}})Li, Zhang, Dubois, Taori, Gulrajani,
  Guestrin, Liang, and Hashimoto}]{alpacaeval}
Xuechen Li, Tianyi Zhang, Yann Dubois, Rohan Taori, Ishaan Gulrajani, Carlos
  Guestrin, Percy Liang, and Tatsunori~B. Hashimoto. 2023{\natexlab{b}}.
\newblock {AlpacaEval: An Automatic Evaluator of Instruction-following Models}.

\bibitem[{Li(2023)}]{li2023open}
Yucheng Li. 2023.
\newblock An open source data contamination report for llama series models.
\newblock \emph{arXiv preprint arXiv:2310.17589}.

\bibitem[{Liang et~al.(2022)Liang, Bommasani, Lee, Tsipras, Soylu, Yasunaga,
  Zhang, Narayanan, Wu, Kumar et~al.}]{liang2022holistic}
Percy Liang, Rishi Bommasani, Tony Lee, Dimitris Tsipras, Dilara Soylu,
  Michihiro Yasunaga, Yian Zhang, Deepak Narayanan, Yuhuai Wu, Ananya Kumar,
  et~al. 2022.
\newblock Holistic evaluation of language models.
\newblock \emph{arXiv preprint arXiv:2211.09110}.

\bibitem[{Lin(2004)}]{lin2004rouge}
Chin-Yew Lin. 2004.
\newblock Rouge: A package for automatic evaluation of summaries.
\newblock In \emph{Text summarization branches out}, pages 74--81.

\bibitem[{Lin et~al.(2021)Lin, Hilton, and Evans}]{lin2021truthfulqa}
Stephanie Lin, Jacob Hilton, and Owain Evans. 2021.
\newblock Truthfulqa: Measuring how models mimic human falsehoods.
\newblock \emph{arXiv preprint arXiv:2109.07958}.

\bibitem[{Lin and Chen(2023)}]{lin2023llmeval}
Yen-Ting Lin and Yun-Nung Chen. 2023.
\newblock Llm-eval: Unified multi-dimensional automatic evaluation for
  open-domain conversations with large language models.
\newblock \emph{arXiv preprint arXiv:2305.13711}.

\bibitem[{Liu et~al.(2023)Liu, Iter, Xu, Wang, Xu, and Zhu}]{liu2023gpteval}
Yang Liu, Dan Iter, Yichong Xu, Shuohang Wang, Ruochen Xu, and Chenguang Zhu.
  2023.
\newblock Gpteval: Nlg evaluation using gpt-4 with better human alignment.
\newblock \emph{arXiv preprint arXiv:2303.16634}.

\bibitem[{Mallio et~al.(2023)Mallio, Sertorio, Bernetti, and
  Beomonte~Zobel}]{mallio2023large}
Carlo~A Mallio, Andrea~C Sertorio, Caterina Bernetti, and Bruno Beomonte~Zobel.
  2023.
\newblock Large language models for structured reporting in radiology:
  performance of gpt-4, chatgpt-3.5, perplexity and bing.
\newblock \emph{La radiologia medica}, pages 1--5.

\bibitem[{MosaicML(2023)}]{mpt7b}
MosaicML. 2023.
\newblock \href {www.mosaicml.com/blog/mpt-7b} {Introducing mpt-7b: A new
  standard for open-source, commercially usable llms}.

\bibitem[{Novikova et~al.(2017)Novikova, Du{\v{s}}ek, Curry, and
  Rieser}]{novikova2017we}
Jekaterina Novikova, Ond{\v{r}}ej Du{\v{s}}ek, Amanda~Cercas Curry, and Verena
  Rieser. 2017.
\newblock Why we need new evaluation metrics for nlg.
\newblock \emph{arXiv preprint arXiv:1707.06875}.

\bibitem[{OpenAI(2023)}]{openai2023gpt4}
OpenAI. 2023.
\newblock \href {http://arxiv.org/abs/2303.08774} {Gpt-4 technical report}.

\bibitem[{Oren et~al.(2023)Oren, Meister, Chatterji, Ladhak, and
  Hashimoto}]{oren2023proving}
Yonatan Oren, Nicole Meister, Niladri Chatterji, Faisal Ladhak, and Tatsunori~B
  Hashimoto. 2023.
\newblock Proving test set contamination in black box language models.
\newblock \emph{arXiv preprint arXiv:2310.17623}.

\bibitem[{Ouyang et~al.(2022)Ouyang, Wu, Jiang, Almeida, Wainwright, Mishkin,
  Zhang, Agarwal, Slama, Ray et~al.}]{ouyang2022training}
Long Ouyang, Jeffrey Wu, Xu~Jiang, Diogo Almeida, Carroll Wainwright, Pamela
  Mishkin, Chong Zhang, Sandhini Agarwal, Katarina Slama, Alex Ray, et~al.
  2022.
\newblock Training language models to follow instructions with human feedback.
\newblock \emph{Advances in Neural Information Processing Systems},
  35:27730--27744.

\bibitem[{Papineni et~al.(2002)Papineni, Roukos, Ward, and
  Zhu}]{papineni2002bleu}
Kishore Papineni, Salim Roukos, Todd Ward, and Wei-Jing Zhu. 2002.
\newblock Bleu: a method for automatic evaluation of machine translation.
\newblock In \emph{Proceedings of the 40th annual meeting of the Association
  for Computational Linguistics}, pages 311--318.

\bibitem[{Peng et~al.(2023)Peng, Li, He, Galley, and Gao}]{peng2023instruction}
Baolin Peng, Chunyuan Li, Pengcheng He, Michel Galley, and Jianfeng Gao. 2023.
\newblock Instruction tuning with gpt-4.
\newblock \emph{arXiv preprint arXiv:2304.03277}.

\bibitem[{Peng et~al.(1997)Peng, Nisbett, and Wong}]{peng1997validity}
Kaiping Peng, Richard~E Nisbett, and Nancy~YC Wong. 1997.
\newblock Validity problems comparing values across cultures and possible
  solutions.
\newblock \emph{Psychological methods}, 2(4):329.

\bibitem[{Pillutla et~al.(2021)Pillutla, Swayamdipta, Zellers, Thickstun,
  Welleck, Choi, and Harchaoui}]{pillutla2021mauve}
Krishna Pillutla, Swabha Swayamdipta, Rowan Zellers, John Thickstun, Sean
  Welleck, Yejin Choi, and Zaid Harchaoui. 2021.
\newblock Mauve: Measuring the gap between neural text and human text using
  divergence frontiers.
\newblock \emph{Advances in Neural Information Processing Systems},
  34:4816--4828.

\bibitem[{Radford et~al.(2018)Radford, Narasimhan, Salimans, Sutskever
  et~al.}]{radford2018improving}
Alec Radford, Karthik Narasimhan, Tim Salimans, Ilya Sutskever, et~al. 2018.
\newblock Improving language understanding by generative pre-training.

\bibitem[{Rajbhandari et~al.(2021)Rajbhandari, Ruwase, Rasley, Smith, and
  He}]{rajbhandari2021zero}
Samyam Rajbhandari, Olatunji Ruwase, Jeff Rasley, Shaden Smith, and Yuxiong He.
  2021.
\newblock Zero-infinity: Breaking the gpu memory wall for extreme scale deep
  learning.
\newblock In \emph{Proceedings of the International Conference for High
  Performance Computing, Networking, Storage and Analysis}, pages 1--14.

\bibitem[{Rasley et~al.(2020)Rasley, Rajbhandari, Ruwase, and
  He}]{rasley2020deepspeed}
Jeff Rasley, Samyam Rajbhandari, Olatunji Ruwase, and Yuxiong He. 2020.
\newblock Deepspeed: System optimizations enable training deep learning models
  with over 100 billion parameters.
\newblock In \emph{Proceedings of the 26th ACM SIGKDD International Conference
  on Knowledge Discovery \& Data Mining}, pages 3505--3506.

\bibitem[{Robinson et~al.(2022)Robinson, Rytting, and
  Wingate}]{Robinson2022LeveragingLL}
Joshua Robinson, Christopher Rytting, and David Wingate. 2022.
\newblock \href {https://api.semanticscholar.org/CorpusID:253098700}
  {Leveraging large language models for multiple choice question answering}.
\newblock \emph{ArXiv}, abs/2210.12353.

\bibitem[{Sainz et~al.(2023)Sainz, Campos, Garc{\'\i}a-Ferrero, Etxaniz,
  de~Lacalle, and Agirre}]{sainz2023nlp}
Oscar Sainz, Jon~Ander Campos, Iker Garc{\'\i}a-Ferrero, Julen Etxaniz,
  Oier~Lopez de~Lacalle, and Eneko Agirre. 2023.
\newblock Nlp evaluation in trouble: On the need to measure llm data
  contamination for each benchmark.
\newblock \emph{arXiv preprint arXiv:2310.18018}.

\bibitem[{Scao et~al.(2022)Scao, Fan, Akiki, Pavlick, Ili{\'c}, Hesslow,
  Castagn{\'e}, Luccioni, Yvon, Gall{\'e} et~al.}]{scao2022bloom}
Teven~Le Scao, Angela Fan, Christopher Akiki, Ellie Pavlick, Suzana Ili{\'c},
  Daniel Hesslow, Roman Castagn{\'e}, Alexandra~Sasha Luccioni, Fran{\c{c}}ois
  Yvon, Matthias Gall{\'e}, et~al. 2022.
\newblock Bloom: A 176b-parameter open-access multilingual language model.
\newblock \emph{arXiv preprint arXiv:2211.05100}.

\bibitem[{Schaeffer(2023)}]{schaeffer2023pretraining}
Rylan Schaeffer. 2023.
\newblock Pretraining on the test set is all you need.
\newblock \emph{arXiv preprint arXiv:2309.08632}.

\bibitem[{Shi et~al.(2023)Shi, Ajith, Xia, Huang, Liu, Blevins, Chen, and
  Zettlemoyer}]{shi2023detecting}
Weijia Shi, Anirudh Ajith, Mengzhou Xia, Yangsibo Huang, Daogao Liu, Terra
  Blevins, Danqi Chen, and Luke Zettlemoyer. 2023.
\newblock Detecting pretraining data from large language models.
\newblock \emph{arXiv preprint arXiv:2310.16789}.

\bibitem[{Srivastava et~al.(2022)Srivastava, Rastogi, Rao, Shoeb, Abid, Fisch,
  Brown, Santoro, Gupta, Garriga-Alonso et~al.}]{srivastava2022beyond}
Aarohi Srivastava, Abhinav Rastogi, Abhishek Rao, Abu Awal~Md Shoeb, Abubakar
  Abid, Adam Fisch, Adam~R Brown, Adam Santoro, Aditya Gupta, Adri{\`a}
  Garriga-Alonso, et~al. 2022.
\newblock Beyond the imitation game: Quantifying and extrapolating the
  capabilities of language models.
\newblock \emph{arXiv preprint arXiv:2206.04615}.

\bibitem[{Sun et~al.(2019)Sun, Qiu, Xu, and Huang}]{sun2019fine}
Chi Sun, Xipeng Qiu, Yige Xu, and Xuanjing Huang. 2019.
\newblock How to fine-tune bert for text classification?
\newblock In \emph{Chinese Computational Linguistics: 18th China National
  Conference, CCL 2019, Kunming, China, October 18--20, 2019, Proceedings 18},
  pages 194--206. Springer.

\bibitem[{Svikhnushina et~al.(2022)Svikhnushina, Filippova, and
  Pu}]{svikhnushina-etal-2022-ieval}
Ekaterina Svikhnushina, Anastasiia Filippova, and Pearl Pu. 2022.
\newblock \href {https://doi.org/10.18653/v1/2022.sigdial-1.41} {i{E}val:
  Interactive evaluation framework for open-domain empathetic chatbots}.
\newblock In \emph{Proceedings of the 23rd Annual Meeting of the Special
  Interest Group on Discourse and Dialogue}, pages 419--431, Edinburgh, UK.
  Association for Computational Linguistics.

\bibitem[{Taori et~al.(2023)Taori, Gulrajani, Zhang, Dubois, Li, Guestrin,
  Liang, and Hashimoto}]{alpaca}
Rohan Taori, Ishaan Gulrajani, Tianyi Zhang, Yann Dubois, Xuechen Li, Carlos
  Guestrin, Percy Liang, and Tatsunori~B. Hashimoto. 2023.
\newblock Stanford alpaca: An instruction-following llama model.
\newblock \url{https://github.com/tatsu-lab/stanford_alpaca}.

\bibitem[{Touvron et~al.(2023{\natexlab{a}})Touvron, Lavril, Izacard, Martinet,
  Lachaux, Lacroix, Rozi{\`e}re, Goyal, Hambro, Azhar
  et~al.}]{touvron2023llama}
Hugo Touvron, Thibaut Lavril, Gautier Izacard, Xavier Martinet, Marie-Anne
  Lachaux, Timoth{\'e}e Lacroix, Baptiste Rozi{\`e}re, Naman Goyal, Eric
  Hambro, Faisal Azhar, et~al. 2023{\natexlab{a}}.
\newblock Llama: Open and efficient foundation language models.
\newblock \emph{arXiv preprint arXiv:2302.13971}.

\bibitem[{Touvron et~al.(2023{\natexlab{b}})Touvron, Martin, Stone, Albert,
  Almahairi, Babaei, Bashlykov, Batra, Bhargava, Bhosale
  et~al.}]{touvron2023llama2}
Hugo Touvron, Louis Martin, Kevin Stone, Peter Albert, Amjad Almahairi, Yasmine
  Babaei, Nikolay Bashlykov, Soumya Batra, Prajjwal Bhargava, Shruti Bhosale,
  et~al. 2023{\natexlab{b}}.
\newblock Llama 2: Open foundation and fine-tuned chat models.
\newblock \emph{arXiv preprint arXiv:2307.09288}.

\bibitem[{Tunstall et~al.(2022)Tunstall, Von~Werra, and
  Wolf}]{tunstall2022natural}
Lewis Tunstall, Leandro Von~Werra, and Thomas Wolf. 2022.
\newblock \emph{Natural language processing with transformers}.
\newblock " O'Reilly Media, Inc.".

\bibitem[{Vaswani et~al.(2017)Vaswani, Shazeer, Parmar, Uszkoreit, Jones,
  Gomez, Kaiser, and Polosukhin}]{vaswani2017attention}
Ashish Vaswani, Noam Shazeer, Niki Parmar, Jakob Uszkoreit, Llion Jones,
  Aidan~N Gomez, {\L}ukasz Kaiser, and Illia Polosukhin. 2017.
\newblock Attention is all you need.
\newblock \emph{Advances in neural information processing systems}, 30.

\bibitem[{Wang et~al.(2018)Wang, Singh, Michael, Hill, Levy, and
  Bowman}]{wangglue}
Alex Wang, Amanpreet Singh, Julian Michael, Felix Hill, Omer Levy, and Samuel~R
  Bowman. 2018.
\newblock Glue: A multi-task benchmark and analysis platform for natural
  language understanding.
\newblock In \emph{International Conference on Learning Representations}.

\bibitem[{Wang et~al.(2023{\natexlab{a}})Wang, Liu, Yue, Tang, Zhang, Jiayang,
  Yao, Gao, Hu, Qi, Wang, Yang, Wang, Xie, Zhang, and Zhang}]{wang2023survey}
Cunxiang Wang, Xiaoze Liu, Yuanhao Yue, Xiangru Tang, Tianhang Zhang, Cheng
  Jiayang, Yunzhi Yao, Wenyang Gao, Xuming Hu, Zehan Qi, Yidong Wang, Linyi
  Yang, Jindong Wang, Xing Xie, Zheng Zhang, and Yue Zhang. 2023{\natexlab{a}}.
\newblock \href {http://arxiv.org/abs/2310.07521} {Survey on factuality in
  large language models: Knowledge, retrieval and domain-specificity}.

\bibitem[{Wang et~al.(2024)Wang, Ning, Pan, Wu, Guo, Deng, Bao, Wang, and
  Zhang}]{wang2024novelqa}
Cunxiang Wang, Ruoxi Ning, Boqi Pan, Tonghui Wu, Qipeng Guo, Cheng Deng,
  Guangsheng Bao, Qian Wang, and Yue Zhang. 2024.
\newblock \href {http://arxiv.org/abs/2403.12766} {Novelqa: A benchmark for
  long-range novel question answering}.

\bibitem[{Wang et~al.(2023{\natexlab{b}})Wang, Li, Chen, Zhu, Lin, Cao, Liu,
  Liu, and Sui}]{wang2023large}
Peiyi Wang, Lei Li, Liang Chen, Dawei Zhu, Binghuai Lin, Yunbo Cao, Qi~Liu,
  Tianyu Liu, and Zhifang Sui. 2023{\natexlab{b}}.
\newblock Large language models are not fair evaluators.
\newblock \emph{arXiv preprint arXiv:2305.17926}.

\bibitem[{Wang et~al.(2023{\natexlab{c}})Wang, Yu, Zeng, Yang, Wang, Chen,
  Jiang, Xie, Wang, Xie et~al.}]{pandalm}
Yidong Wang, Zhuohao Yu, Zhengran Zeng, Linyi Yang, Cunxiang Wang, Hao Chen,
  Chaoya Jiang, Rui Xie, Jindong Wang, Xing Xie, et~al. 2023{\natexlab{c}}.
\newblock Pandalm: An automatic evaluation benchmark for llm instruction tuning
  optimization.
\newblock \emph{arXiv preprint arXiv:2306.05087}.

\bibitem[{Wang et~al.(2022)Wang, Kordi, Mishra, Liu, Smith, Khashabi, and
  Hajishirzi}]{wang2022self}
Yizhong Wang, Yeganeh Kordi, Swaroop Mishra, Alisa Liu, Noah~A Smith, Daniel
  Khashabi, and Hannaneh Hajishirzi. 2022.
\newblock Self-instruct: Aligning language model with self generated
  instructions.
\newblock \emph{arXiv preprint arXiv:2212.10560}.

\bibitem[{Wei et~al.(2023)Wei, Zhao, Zhang, Zhu, Wang, Yang, Li, Cheng, L{\"u},
  Hu et~al.}]{wei2023skywork}
Tianwen Wei, Liang Zhao, Lichang Zhang, Bo~Zhu, Lijie Wang, Haihua Yang, Biye
  Li, Cheng Cheng, Weiwei L{\"u}, Rui Hu, et~al. 2023.
\newblock Skywork: A more open bilingual foundation model.
\newblock \emph{arXiv preprint arXiv:2310.19341}.

\bibitem[{Wolf et~al.(2019)Wolf, Debut, Sanh, Chaumond, Delangue, Moi, Cistac,
  Rault, Louf, Funtowicz et~al.}]{wolf2019huggingface}
Thomas Wolf, Lysandre Debut, Victor Sanh, Julien Chaumond, Clement Delangue,
  Anthony Moi, Pierric Cistac, Tim Rault, R{\'e}mi Louf, Morgan Funtowicz,
  et~al. 2019.
\newblock Huggingface's transformers: State-of-the-art natural language
  processing.
\newblock \emph{arXiv preprint arXiv:1910.03771}.

\bibitem[{Workshop et~al.(2022)Workshop, Scao, Fan, Akiki, Pavlick, Ili{\'c},
  Hesslow, Castagn{\'e}, Luccioni, Yvon et~al.}]{workshop2022bloom}
BigScience Workshop, Teven~Le Scao, Angela Fan, Christopher Akiki, Ellie
  Pavlick, Suzana Ili{\'c}, Daniel Hesslow, Roman Castagn{\'e}, Alexandra~Sasha
  Luccioni, Fran{\c{c}}ois Yvon, et~al. 2022.
\newblock Bloom: A 176b-parameter open-access multilingual language model.
\newblock \emph{arXiv preprint arXiv:2211.05100}.

\bibitem[{Xie et~al.(2024)Xie, Zeng, Yu, Gao, Zhang, and Ye}]{xie2024codeshell}
Rui Xie, Zhengran Zeng, Zhuohao Yu, Chang Gao, Shikun Zhang, and Wei Ye. 2024.
\newblock \href {http://arxiv.org/abs/2403.15747} {Codeshell technical report}.

\bibitem[{Xu et~al.(2020)Xu, Hu, Zhang, Li, Cao, Li, Xu, Sun, Yu, Yu
  et~al.}]{xu2020clue}
Liang Xu, Hai Hu, Xuanwei Zhang, Lu~Li, Chenjie Cao, Yudong Li, Yechen Xu, Kai
  Sun, Dian Yu, Cong Yu, et~al. 2020.
\newblock Clue: A chinese language understanding evaluation benchmark.
\newblock In \emph{Proceedings of the 28th International Conference on
  Computational Linguistics}, pages 4762--4772.

\bibitem[{Yang et~al.(2022)Yang, Zhang, Qin, Li, Wang, Liu, Wang, Xie, and
  Zhang}]{yang2022glue}
Linyi Yang, Shuibai Zhang, Libo Qin, Yafu Li, Yidong Wang, Hanmeng Liu, Jindong
  Wang, Xing Xie, and Yue Zhang. 2022.
\newblock Glue-x: Evaluating natural language understanding models from an
  out-of-distribution generalization perspective.
\newblock \emph{arXiv preprint arXiv:2211.08073}.

\bibitem[{Yang et~al.(2023)Yang, Zhang, Yu, Bao, Wang, Wang, Xu, Ye, Xie, Chen,
  and Zhang}]{yang2023supervised}
Linyi Yang, Shuibai Zhang, Zhuohao Yu, Guangsheng Bao, Yidong Wang, Jindong
  Wang, Ruochen Xu, Wei Ye, Xing Xie, Weizhu Chen, and Yue Zhang. 2023.
\newblock \href {http://arxiv.org/abs/2312.15918} {Supervised knowledge makes
  large language models better in-context learners}.

\bibitem[{Yoo et~al.(2003)Yoo, Jette, and Grondona}]{yoo2003slurm}
Andy~B Yoo, Morris~A Jette, and Mark Grondona. 2003.
\newblock Slurm: Simple linux utility for resource management.
\newblock In \emph{Workshop on job scheduling strategies for parallel
  processing}, pages 44--60. Springer.

\bibitem[{Yu et~al.(2024)Yu, Gao, Yao, Wang, Ye, Wang, Xie, Zhang, and
  Zhang}]{yu2024kieval}
Zhuohao Yu, Chang Gao, Wenjin Yao, Yidong Wang, Wei Ye, Jindong Wang, Xing Xie,
  Yue Zhang, and Shikun Zhang. 2024.
\newblock Kieval: A knowledge-grounded interactive evaluation framework for
  large language models.
\newblock \emph{arXiv preprint arXiv:2402.15043}.

\bibitem[{Yuan et~al.(2021)Yuan, Neubig, and Liu}]{yuan2021bartscore}
Weizhe Yuan, Graham Neubig, and Pengfei Liu. 2021.
\newblock Bartscore: Evaluating generated text as text generation.
\newblock \emph{Advances in Neural Information Processing Systems},
  34:27263--27277.

\bibitem[{Zellers et~al.(2019)Zellers, Holtzman, Bisk, Farhadi, and
  Choi}]{zellers2019hellaswag}
Rowan Zellers, Ari Holtzman, Yonatan Bisk, Ali Farhadi, and Yejin Choi. 2019.
\newblock Hellaswag: Can a machine really finish your sentence?
\newblock \emph{arXiv preprint arXiv:1905.07830}.

\bibitem[{Zeng et~al.(2022)Zeng, Liu, Du, Wang, Lai, Ding, Yang, Xu, Zheng, Xia
  et~al.}]{zeng2022glm}
Aohan Zeng, Xiao Liu, Zhengxiao Du, Zihan Wang, Hanyu Lai, Ming Ding, Zhuoyi
  Yang, Yifan Xu, Wendi Zheng, Xiao Xia, et~al. 2022.
\newblock Glm-130b: An open bilingual pre-trained model.
\newblock \emph{arXiv preprint arXiv:2210.02414}.

\bibitem[{Zeng et~al.(2024)Zeng, Wang, Xie, Ye, and Zhang}]{zeng2024coderujb}
Zhengran Zeng, Yidong Wang, Rui Xie, Wei Ye, and Shikun Zhang. 2024.
\newblock \href {http://arxiv.org/abs/2403.19287} {Coderujb: An executable and
  unified java benchmark for practical programming scenarios}.

\bibitem[{Zeng et~al.(2023)Zeng, Yu, Gao, Meng, Goyal, and
  Chen}]{zeng2023evaluating}
Zhiyuan Zeng, Jiatong Yu, Tianyu Gao, Yu~Meng, Tanya Goyal, and Danqi Chen.
  2023.
\newblock Evaluating large language models at evaluating instruction following.
\newblock \emph{arXiv preprint arXiv:2310.07641}.

\bibitem[{Zhang et~al.(2022)Zhang, Roller, Goyal, Artetxe, Chen, Chen, Dewan,
  Diab, Li, Lin et~al.}]{zhang2022opt}
Susan Zhang, Stephen Roller, Naman Goyal, Mikel Artetxe, Moya Chen, Shuohui
  Chen, Christopher Dewan, Mona Diab, Xian Li, Xi~Victoria Lin, et~al. 2022.
\newblock Opt: Open pre-trained transformer language models.
\newblock \emph{arXiv preprint arXiv:2205.01068}.

\bibitem[{Zhang et~al.(2019)Zhang, Kishore, Wu, Weinberger, and
  Artzi}]{zhang2019bertscore}
Tianyi Zhang, Varsha Kishore, Felix Wu, Kilian~Q Weinberger, and Yoav Artzi.
  2019.
\newblock Bertscore: Evaluating text generation with bert.
\newblock \emph{arXiv preprint arXiv:1904.09675}.

\bibitem[{Zheng et~al.(2023{\natexlab{a}})Zheng, Zhou, Meng, Zhou, and
  Huang}]{zheng2023large}
Chujie Zheng, Hao Zhou, Fandong Meng, Jie Zhou, and Minlie Huang.
  2023{\natexlab{a}}.
\newblock Large language models are not robust multiple choice selectors.
\newblock In \emph{The Twelfth International Conference on Learning
  Representations}.

\bibitem[{Zheng et~al.(2023{\natexlab{b}})Zheng, Chiang, Sheng, Zhuang, Wu,
  Zhuang, Lin, Li, Li, Xing et~al.}]{mtbench}
Lianmin Zheng, Wei-Lin Chiang, Ying Sheng, Siyuan Zhuang, Zhanghao Wu, Yonghao
  Zhuang, Zi~Lin, Zhuohan Li, Dacheng Li, Eric Xing, et~al. 2023{\natexlab{b}}.
\newblock Judging llm-as-a-judge with mt-bench and chatbot arena.
\newblock \emph{arXiv preprint arXiv:2306.05685}.

\bibitem[{Zhou et~al.(2023)Zhou, Zhu, Chen, Chen, Zhao, Chen, Lin, Wen, and
  Han}]{zhou2023dont}
Kun Zhou, Yutao Zhu, Zhipeng Chen, Wentong Chen, Wayne~Xin Zhao, Xu~Chen,
  Yankai Lin, Ji-Rong Wen, and Jiawei Han. 2023.
\newblock Don't make your llm an evaluation benchmark cheater.
\newblock \emph{arXiv preprint arXiv:2311.01964}.

\bibitem[{Zhu et~al.(2023{\natexlab{a}})Zhu, Chen, Wang, Gong, Yang, and
  Xie}]{zhu2023dyval}
Kaijie Zhu, Jiaao Chen, Jindong Wang, Neil~Zhenqiang Gong, Diyi Yang, and Xing
  Xie. 2023{\natexlab{a}}.
\newblock Dyval: Graph-informed dynamic evaluation of large language models.
\newblock \emph{arXiv preprint arXiv:2309.17167}.

\bibitem[{Zhu et~al.(2023{\natexlab{b}})Zhu, Wang, Zhou, Wang, Chen, Wang,
  Yang, Ye, Gong, Zhang et~al.}]{zhu2023promptbench}
Kaijie Zhu, Jindong Wang, Jiaheng Zhou, Zichen Wang, Hao Chen, Yidong Wang,
  Linyi Yang, Wei Ye, Neil~Zhenqiang Gong, Yue Zhang, et~al.
  2023{\natexlab{b}}.
\newblock Promptbench: Towards evaluating the robustness of large language
  models on adversarial prompts.
\newblock \emph{arXiv preprint arXiv:2306.04528}.

\end{thebibliography}

\label{sec:ethical-consideration}


\end{document}